\pdfoutput=1
\documentclass[10pt,twocolumn,letterpaper]{article}

\usepackage{iccv}              

\usepackage[accsupp]{axessibility}
\usepackage{amsmath}
\usepackage{afterpage}
\usepackage{amssymb}
\usepackage{booktabs}
\usepackage{pifont}
\usepackage{color}
\usepackage[table]{xcolor}
\usepackage{bm}
\usepackage{multirow}
\usepackage{soul}
\usepackage{bbm}
\usepackage{subcaption}
\usepackage{xurl}
\usepackage{graphicx}
\usepackage{tabularx}  
\usepackage{listings}
\lstdefinestyle{promptstyle}{
    basicstyle=\ttfamily\footnotesize,
    backgroundcolor=\color{gray!10},
    frame=single,
    breaklines=true}

\definecolor{citecolor}{HTML}{0071bc}
\usepackage[pagebackref=true,breaklinks=true,colorlinks,citecolor=citecolor,bookmarks=false]{hyperref}

\def\ours{\texttt{GLEVR}}

\def\paperID{9204} 
\def\confName{ICCV}
\def\confYear{2025}

\title{Keystep Recognition using Graph Neural Networks}

\author{Julia Lee Romero$^{1}$ 
\quad Kyle Min$^{2}$ \quad Subarna Tripathi$^{2}$ \quad Morteza Karimzadeh$^{1}$\\
$^{1}$University of Colorado Boulder \quad\quad $^{2}$Intel Labs\\
{\tt\small \{julia.romero, karimzadeh\}@colorado.edu, \{kyle.min, subarna.tripathi\}@intel.com}
}

\begin{document}
\maketitle
\begin{abstract}

We pose keystep recognition as a node classification task, and propose a flexible graph-learning framework for fine-grained keystep recognition that is able to effectively leverage long-term dependencies in egocentric videos. Our approach, termed GLEVR, consists of constructing a graph where each video clip of the egocentric video corresponds to a node. The constructed graphs are sparse and computationally efficient, outperforming existing larger models substantially. We further leverage alignment between egocentric and exocentric videos during training for improved inference on egocentric videos, as well as adding automatic captioning as an additional modality. We consider each clip of each exocentric video (if available) or video captions as additional nodes during training. We examine several strategies to define connections across these nodes. We perform extensive experiments on the Ego-Exo4D dataset and show that our proposed flexible graph-based framework notably outperforms existing methods.



\end{abstract}    

\section{Introduction}

\begin{figure}[!t]
  \centering
  \begin{subfigure}{0.99\linewidth}
    \centering
    \includegraphics[width=\linewidth]{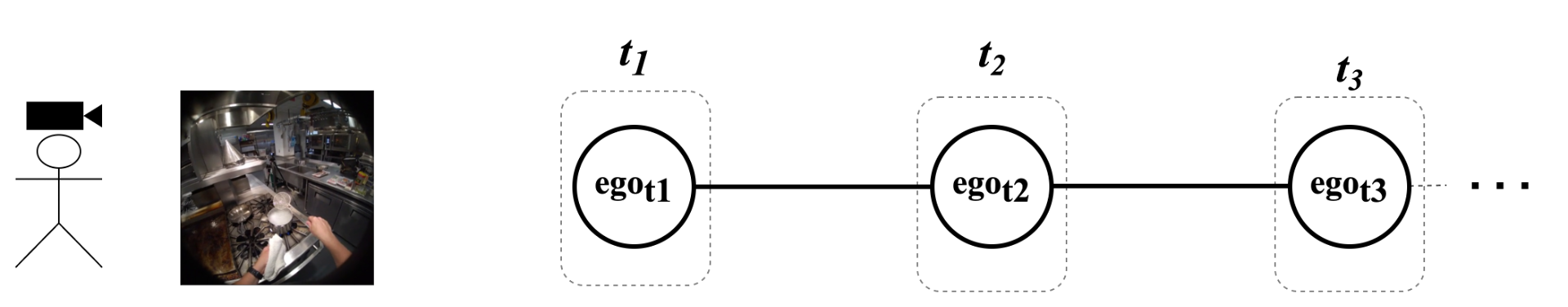}
    \caption{Egocentric Vision Graph}
    \label{fig:multiview1}
    \vspace{-8pt}
  \end{subfigure}
  \hfill
  \vspace{0.4cm} 
  \begin{subfigure}{0.99\linewidth}
    \centering
    \includegraphics[width=\linewidth]{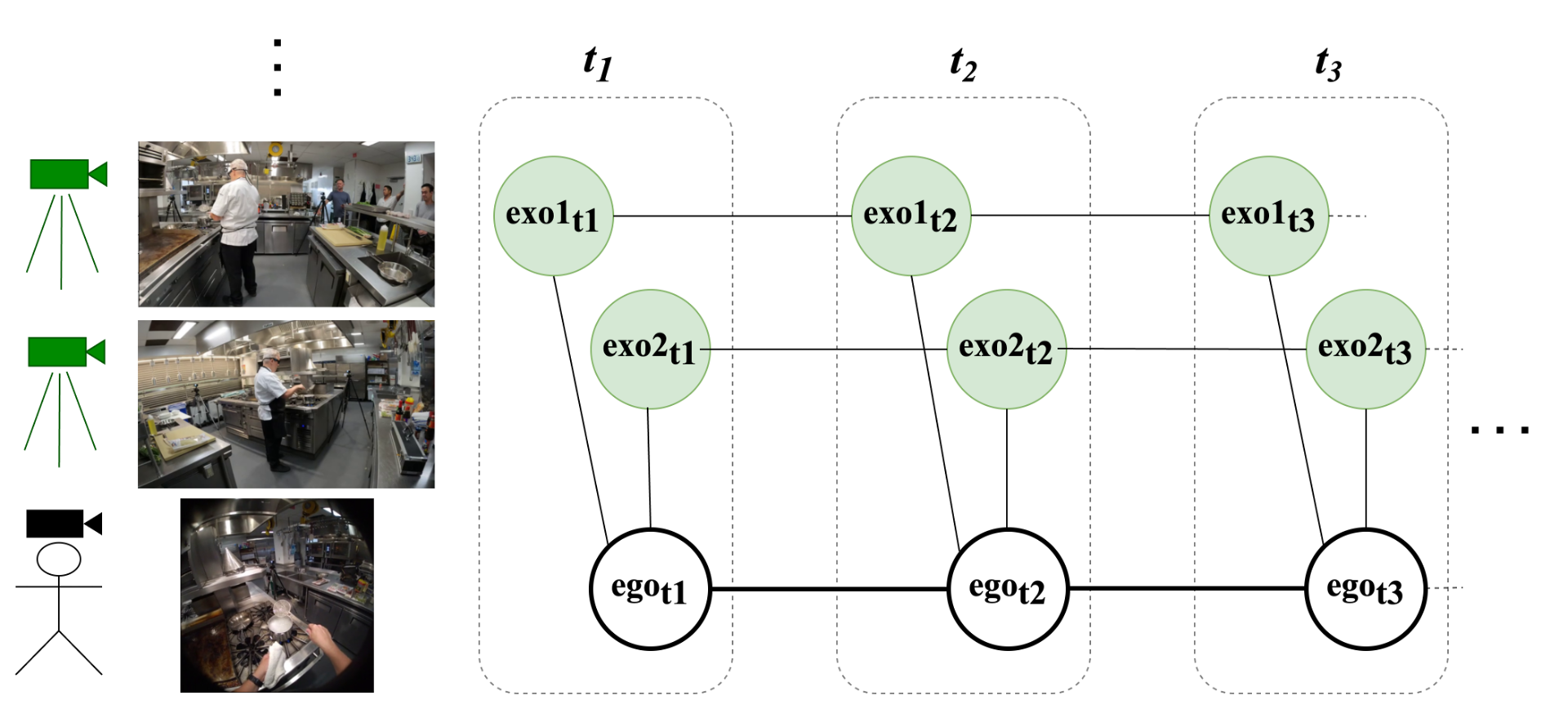}
    \caption{Multi-view Vision Graph}
    \label{fig:multiview2}
    \vspace{-6pt}
  \end{subfigure}
  \begin{subfigure}{0.99\linewidth}
    \centering
    \includegraphics[width=\linewidth]{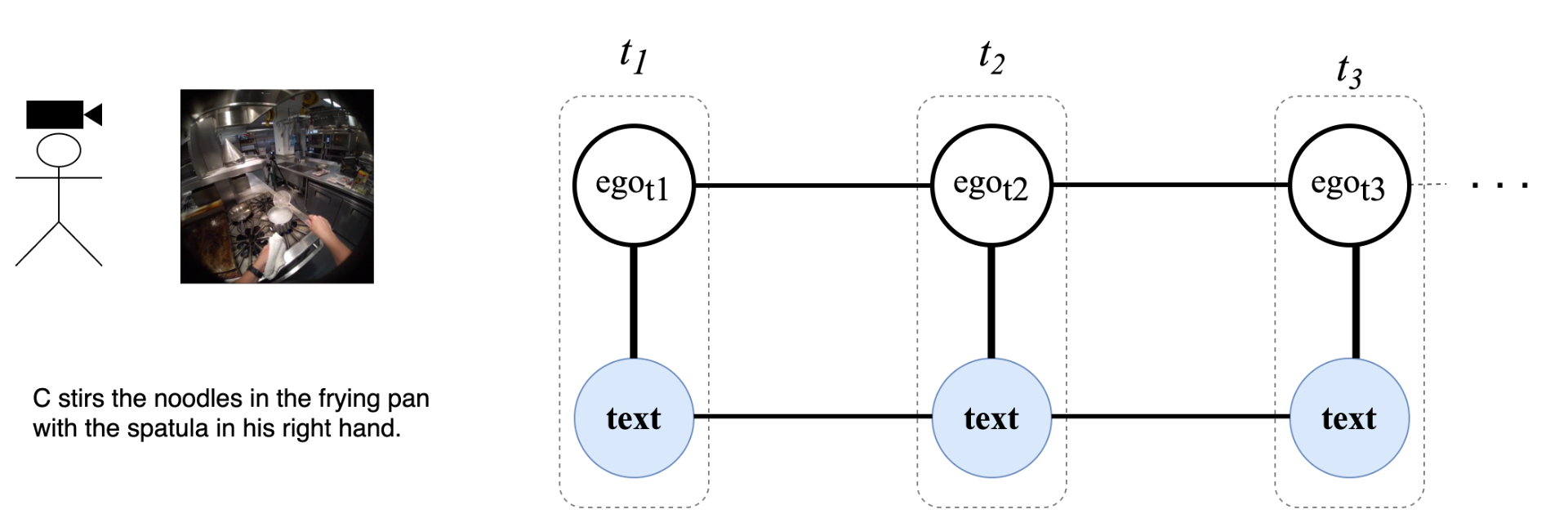}
    \caption{Egocentric Multimodal Graph}
    \label{fig:ego-hetero}
    \vspace{-6pt}
  \end{subfigure}
  \caption{\ours{}: Graph-based representation learning for keystep recognition. Learning leverages several strategies such as utilizing additional exocentric views via multi-view alignment, and incorporating complimentary information from other modalities. The inference is always over the egocentric view only.}
  \label{fig:GLEVR}
     \vspace{-8pt}
\end{figure}



We develop a compute-efficient approach for fine-grained keystep recognition on procedural egocentric videos that leverages long-term dependencies more efficiently and improves egocentric-only predictive performance. Our framework also allows for further improvement by incorporating a variable number of exocentric videos available only during training,
and optionally incorporating automatic captioning as an additional input modality
as demonstrated in Figure \ref{fig:GLEVR}.





To summarize, below is the list of our contributions.  
\begin{itemize}
    \setlength{\itemsep}{1pt}
    \item {\bf Graph-based representation learning for long-videos:} 
    We propose a compute-efficient, graph-based representation learning framework for modeling temporal dynamics in long-form videos. We formulate keystep recognition as a node classification problem. We refer to this framework as \ours{}, Graph Learning on Egocentric Videos for keystep Recognition. 
    \item {\bf Multi-view alignment with graph:} 
    \ours{} fuses complementary information from a variable number of views by learning multi-view alignment. 
    Leveraging a variable number of views (if present) during training as \emph{one sample} makes \ours{} stand out from the baselines, which treat them as separate samples. 

    \item {\bf Heterogenous graph learning using multimodal alignment:} We also present a study on how to leverage complementary multimodal information using \ours{}-Hetero framework. We observe that \ours{}-hetero with video narrations improves accuracy compared to its visual-only counterpart. 
    \item {\bf Extensive experiments on Ego-Exo4D:} 
    \ours{} notably outperforms existing 
    methods by more than $16\%$ and $12\%$, on the validation and test set, respectively.
\end{itemize}



\section{Related Work}

\noindent\textbf{\textbf{Keystep Recognition on the Ego-Exo4D Dataset}}
Prior work on this task evaluates a diverse set of baseline approaches, such as models learned for action classification (TimesFormer~\cite{Timesformer_ICML_21}), video-language pre-training (EgoVLP2~\cite{egovlpv2}), view-invariant two-stage training, view-point distillation and Ego-Exo transfer~\cite{EgoExo_ICCV21} with an improved backbone. Using both egocentric and exocentric videos during training provides similar or even worse performance when compared to the already-low metrics on Ego-view only in the first two baselines, indicating that a more advanced method is required to fully leverage these multi-view data.

\noindent\textbf{\textbf{Graph-based Representations for Video Understanding}}
A line of work~\cite{zhao2023constructing,arnab2021unified,rai2021home} explored scene graphs for video understanding, emphasizing the effectiveness of the structured representations in understanding temporal actions and interactions within videos. It has also been shown that graph-based representations without ground-truth graph annotations can be effective for lightweight frameworks for video understanding applications, including active speaker detection~\cite{spell2022,minintel}. In comparison, graph-based representation learning for egocentric videos is a relatively nascent field. Recently, the Ego4D~\cite{grauman2022ego4d} dataset has motivated a few works focusing on egocentric videos. In~\cite{min2022intel, min2023sthg}, the authors show how the graph-based representation can be leveraged for audio-video diarization in egocentric videos. The work of \cite{rodin2024action} introduces a temporally evolving graph structure of the actions performed in egocentric videos and proposes a new graph generation task.

Our approach is distinguished from the literature in the sense that we formulate the challenging problem of fine-grained keystep recognition as node classification on a 
graph constructed from the input egocentric videos, 
while leveraging the variable number of exocentric videos only during training.
\section{Experiments}
\subsection{Problem Formulation}
Ego-Exo4D's keystep recognition task is classification on trimmed video clips \cite{egoexo4d}. At training time, we are given egocentric videos and their aligned exocentric videos, while at test time, only the egocentric videos and the trimmed clip segments are provided. 

\subsection{Graph Architecture}
 We investigate components of the graph architecture, including edge connections, context-length for long-form reasoning, ability to leverage ego-exo relations, and incorporation of multimodal data, specifically text descriptions of segments.


We construct a graph given the egocentric input video and keystep segments within the video. Each node corresponds to a keystep segment, and temporal edges connect subsequent node segments. The model is trained on a node classification task, such that a keystep prediction is made for each node in the graph. In this setup, there is only one node type (vision) and one edge type (temporal).

Building on the egocentric vision graph, we define a multiview graph, which adds a node for each
exocentric view for each keystep segment, drawing ego-exo edges between nodes of corresponding segments.

The graph framework also lets us utilize complementary multimodal information such as 
video narrations.
We fine-tune the recent VideoRecap \cite{islam2024video} framework on Ego-Exo4D to generate text narrations for 4-second clips spanning the video length and summarize these narrations for keystep segments using LlaMA-3-8B fine-tuned for summarization \cite{song2024learning}. We extracted \emph{LongCLIP}\cite{zhang2024long} features from the generated segment-level captions.

All prior approaches operate directly on video while \ours{} uses pre-extracted visual features, so we evaluate a simple multi-layer perceptron (MLP) on these visual features as an additional baseline comparison. We use the pre-extracted Omnivore Swin-L vision features on each video ~\cite{Omnivor_2022_CVPR}, provided by  Ego-Exo4D.

\subsection{Experimental Setup}
\textbf{Dataset}
We use the full Ego-Exo4D dataset for the fine-grained keystep recognition task \cite{egoexo4d} in this work. This includes 1088 videos taking a total of 87 hours, with an average of $4.8$ minutes per video, and 289 classes.

In our graph setup, the number of graphs in the ego-only and the multi-view setups are equal since all available views are used in one graph. Each node in the graph is used during training, so the multi-view setup has more samples than the ego-only setup. 

\textbf{Training and Evaluation}
For training and evaluation, we employ cross-validation and split the graphs into five splits. We report the validation performance as the average accuracy on the five splits. The framework is trained on each set of 4 splits and evaluated on the remaining split. In accordance with the Ego-Exo4D keystep recognition task definition, we evaluate top-1 keystep recognition accuracy and also report an F1 score at the threshold of 0.1 (F1@0.1). 



\subsection{Experimental Results}

\begin{table}[h]
    \centering
    \setlength{\tabcolsep}{9pt}
    \resizebox{\linewidth}{!}{
    \begin{tabular}{l|cc|cc}
        \hline
        \textbf{Method} & \textbf{Narration} & \textbf{Val Acc} & \textbf{Test Acc} \\
        \hline
        TimeSFormer~\cite{Timesformer_ICML_21} (K600) & \ding{55} & 35.25 & 35.93 \\
        EgoVLPv2~\cite{egovlpv2} (EgoExo) & \ding{55} & 38.21 & 38.69 \\
        VI Encoder~\cite{VI-encoder-2018} (EgoExo) & \ding{55} & 40.23 & 41.53 \\
        Viewpoint Distillation~\cite{viewppoint-distillation-hinton2015} & \ding{55} & 37.79 & 39.49 \\
        Ego-Exo Transfer MAE~\cite{ego-exo-2021} & \ding{55} & 36.71 & 35.57 \\
        \midrule
        MLP baseline & \ding{55} & 40.40 & - \\
        \textbf{\ours{}} & \ding{55} & 54.69 & 52.36 \\
        \midrule
        \textbf{\ours-Hetero} & \checkmark & \textbf{56.99}  & \textbf{53.65} \\
        \bottomrule
    \end{tabular} 
    }
    \vspace{-7pt}
    \caption{Top-1 accuracy on keystep recognition. MLP baseline and our model use frame-level Omnivore features.
 }
    \label{baselines}
\end{table}

\subsubsection{Comparison to Existing Approaches}

\textbf{Performance}
We report the results of prior approaches alongside \ours{}'s best results and an MLP baseline on omnivore features in Table ~\ref{baselines}. 
A strong MLP baseline already outperforms existing methods, showing the effectiveness of the visual features. 
These results demonstrate that \ours{} improves upon prior methods by substantial margins. 
First, we compare the performance of \ours{} with visual features input only. 
\ours{} outperforms the present state-of-the-art method, VI Encoder, by 16.76 and 12.12 points in accuracy on validation and test sets, respectively. 
\par
Next, we show how leveraging off-the-shelf narration generation models can improve keystep recognition further. 
Our framework is able to leverage automatically generated caption features in addition to the visual features and shows $1.29\%$ additional improvement on the test set compared to the vision-only model.  
\par
\textbf{Ability to leverage long-form reasoning} We evaluate \ours{} on varying temporal context lengths to demonstrate its capacity for long-form reasoning. Table~\ref{table:temporal-context-size} shows that increasing the temporal context significantly improves key-step recognition.
When context is limited to a quarter of the video (short context), performance drops compared to using the full video.  On average, short-context graphs contain 5.63 segments, while full-context graphs contain 22.52. As shown in Table~\ref{table:temporal-context-size}, performance improves by over 14 points (Acc) when moving from no context to full context. Existing methods based on dense computation can not handle longer temporal context \cite{egovlpv2, Timesformer_ICML_21}. \ours{} can easily process and reason over the whole video, due to the sparsity of the underlying graph architecture.

\begin{table}[!t]
    \centering
        \setlength{\tabcolsep}{26pt}
        \resizebox{\linewidth}{!}{
        \begin{tabular}{l|cc}
        \toprule
        Context size
           & Acc ($\uparrow$) & F1@0.1 ($\uparrow$) \\
           \midrule
        no-context &  40.40 & 43.00 \\
        short context  & 51.18 & 44.67  \\
        full context  & 54.69 & 52.36 \\
          \bottomrule
        \end{tabular}
        }
        \vspace{-7pt}
        \captionof{table}{Results of varying the context length of the Ego-only setting. No context means each segment is treated independently, this is the MLP baseline case. Short and full context denote quarter duration and full duration of the video at a time, respectively.   
        }
        \label{table:temporal-context-size}
\end{table}

\par
\textbf{Ability to leverage multiple views} 
One of the primary motivations behind creating Ego-Exo4D dataset is to define and address the core research challenges in the domain of egocentric perception of skilled activity, particularly when \emph{ego-exo data is available for training}, but not during training. Leveraging a variable number of views (if present) during training as one sample makes \ours{} stand out from the baselines, which treat them as separate samples.

\begin{table}[h]
    \centering
    \setlength{\tabcolsep}{9pt}
    \resizebox{\linewidth}{!}{
    \begin{tabular}{l|cc|cc}
        \hline
        \textbf{Method} & \textbf{Val Acc} & \textbf{Val gain} & \textbf{Test Acc} & \textbf{Test Gain}\\
        \hline
         TimeSFormer~\cite{Timesformer_ICML_21} (K600) & 32.67 & -2.58 & 29.84 &-5.40 \\
        EgoVLPv2~\cite{egovlpv2} (Ego4D) & 37.03 &0.14 & 36.84 &-0.67\\
        EgoVLPv2~\cite{egovlpv2} (EgoExo) & 38.21 & 0.60 & 38.69 &0.84\\
        VI Encoder~\cite{VI-encoder-2018} (EgoExo) & 40.23 & - & 40.61 & -\\
        Viewpoint Distillation~\cite{viewppoint-distillation-hinton2015} & 37.79 & - & 38.10 &-\\
        Ego-Exo Transfer MAE~\cite{ego-exo-2021} & 36.71 &-& 35.57 &- \\
        \midrule
        \textbf{\ours} & \textbf{56.74} & \textbf{2.05} & \textbf{53.08} & 0.72\\
        \bottomrule
    \end{tabular} 
    }
    \vspace{-7pt}
    \caption{Performance gain from egocentric-only to multiview. 
 }
    \label{tab:multi-view alignment}
\end{table}

Ego-Exo4D~\cite{ego-exo-2021} observed and emphasized that different approaches respond differently to the addition of exo-view videos during training. 
Many of the existing approaches such as TimeSFormer (K600)~\cite{Timesformer_ICML_21} suffer a drop in performance when exo views are added to the training. 

\par
\ours{} exhibits a performance gain of $2\%$ from the ego-only to the multi-view setup as shown in the Table~\ref{tab:multi-view alignment}. 
Our results demonstrate that \ours{} is able to effectively leverage multi-view information without increasing the sample size.
While Viewpoint Distillation~\cite{viewppoint-distillation-hinton2015} and \ours{} utilize all ego-exo views simultaneously in a multi-view setup, \ours{} clearly outperforms ~\cite{viewppoint-distillation-hinton2015} by $19\%$. 

\par
\textbf{Model Efficiency}
\ours{} is a lightweight and compute-efficient framework (Table ~\ref{table:model-efficiency}). It can be efficiently trained from randomly initialized weights in a single phase. Another key distinction is that \ours{} can use any pre-extracted frame-level visual features, while existing methods operate on video. This significantly reduces the computational burden associated with processing high-dimensional video data, accelerating the training and development process and lowering hardware requirements. The memory requirements scale with the number of segments in the video, since there is a node for each segment.

\begin{table}[!t]
    \centering
        \setlength{\tabcolsep}{15pt}
        \resizebox{0.9\linewidth}{!}{
        \begin{tabular}{l|c}
        \toprule
        \multirow{1}{*}{Model} & Model Size (MB) \\ \midrule
        \ours{} graph only   & 91 \\ 
        \ours{} Hetero graph only  & 173 \\
        \ours{} + Omnivore Swin-L  & 539 \\
        TimeSFormer & 929 \\ 
        EgoVLPv2  & 4300 \\
   
        \bottomrule
        \end{tabular}
        }
        \vspace{-7pt}
        \caption{Comparison of model sizes.}
        \label{table:model-efficiency}
\end{table}





\subsubsection{Study on leveraging multimodal information}

\begin{table}[!t]
    \centering
        \setlength{\tabcolsep}{26pt}
        \resizebox{\linewidth}{!}{
        \begin{tabular}{l|cc}
        \toprule
        Features
           & Acc & F1@0.1 
           \\ \midrule
        \ours{} &  54.69 & 52.36 \\
        \ours{}-Hetero  & 56.99 & 53.65 \\
          \bottomrule
        \end{tabular}
        }
        \vspace{-7pt}
        \captionof{table}{Results of multimodal experiment with \ours{} on visual features only and \ours{}-Hetero with \emph{LongCLIP} segment narration features for the Ego-only configuration.  
        }
        \label{table:cross-view-alignment}
\end{table}


The flexible graph framework is amenable for utilizing complementary multimodal information in a heterogeneous graph learning framework as illustrated in Fig. \ref{fig:GLEVR}.

\textbf{Utilizing narrations}
We hypothesize that narrations aid fine-grained keystep recognition and demonstrate that our framework can effectively leverage them. Using \emph{LongCLIP} features from ground truth manual narrations in a heterogeneous graph yields $94.87\%$ accuracy on the validation set, and we consider this as the upper bound or golden reference. 


To approach this golden reference without ground truth narrations, we fine-tune a state-of-the-art video narration model, VideoRecap~\cite{islam2024video} on Ego-Exo4D to generate clip-level captions. We then derive segment-level narrations in two ways: (1) by concatenating all clip-level captions, and (2) by summarizing them using a large language model, LlaMA-3 \cite{song2024learning, grattafiori2024llama}. 
Extracted \emph{LongCLIP} features are used as text node features in \ours{}-Hetero. 

We find that simple concatenation slightly degrades performance, while LLM-generated summaries improve accuracy over visual-only models (Table~\ref{table:narration-experiments}).

\begin{table}[!t]
    \centering
        \setlength{\tabcolsep}{16pt}
        \resizebox{\linewidth}{!}{
        \begin{tabular}{l|cc}
        \toprule 
        Narration types        
           & Acc & F1@0.1 
           \\ \toprule
        GT narration* &  94.87* & 91.05* \\
        GT atomic action description  & 65.38 & 65.17 \\
        \midrule
        Concat clip narrations  & 54.96 & 55.74 \\
        Segment-Level Summary  & 56.45 & 56.21 \\
          \bottomrule
        \end{tabular}
        }
        \vspace{-7pt}
        \captionof{table}{Results for different text annotations with \ours{}-Hetero for the multiview configuration. * denotes golden reference or upper bound with ground truth narrations.  
        }
        \label{table:narration-experiments}
\end{table}

\section{Conclusion and Discussions}
We presented \ours{}, a graph learning framework for efficient fine-grained keystep recognition over long videos. We posed the keystep recognition problem as a node classification problem in the constructed graphs. We showed that our framework effectively performs on long videos, and is able to leverage complementary information from multiple views that are available only during training. 
We considered each clip per viewpoint as a node in a graph and experimented with several different ways of connecting those nodes.
Our experiments on Ego-Exo4D validation and test set showed that \ours{} notably outperforms existing methods as measured by the accuracy and F-1 score metrics. 
\ours{} emerged as a new state-of-the-art method for the fine-grained keystep recognition task, all the while being memory and compute-efficient.
We further proposed a heterogeneous graph learning framework, \ours{}-Hetero, to leverage multimodal data extracted from \emph{generated} video narrations and show further performance improvement.  
We believe this work will motivate a line of follow-up research dedicated to memory- and compute-efficient video understanding.  


{
    \small
    \bibliographystyle{ieeenat_fullname}
    \bibliography{egbib}
}

\end{document}